\documentclass[sigconf]{acmart}

\AtBeginDocument{%
  }

\setcopyright{acmlicensed}
\copyrightyear{2018}
\acmYear{2018}
\acmDOI{XXXXXXX.XXXXXXX}

\acmConference[KDD Cup 24]{}{August 25--29}{Barcelona, Spain}
\acmISBN{978-1-4503-XXXX-X/18/06}

\begin{document}

\title{MARAGS: A Multi-Adapter System for Multi-Task Retrieval Augmented Generation Question Answering}

\author{Mitchell DeHaven}
\email{mdehaven@darkhive.com}
\affiliation{%
  \institution{Darkhive}
  \city{San Antonio}
  \state{Texas}
  \country{USA}
}

\begin{abstract}

  In this paper we present a multi-adapter retrieval augmented generation system (MARAGS) for Meta's Comprehensive RAG (CRAG) competition for KDD CUP 2024. CRAG is a question answering dataset contains 3 different subtasks aimed at realistic question and answering RAG related tasks, with a diverse set of question topics, question types, time dynamic answers, and questions featuring entities of varying popularity.
  
  Our system follows a standard setup for web based RAG, which uses processed web pages to provide context for an LLM to produce generations, while also querying API endpoints for additional information. MARAGS also utilizes multiple different adapters to solve the various requirements for these tasks with a standard cross-encoder model for ranking candidate passages relevant for answering the question. Our system achieved 2nd place for Task 1 as well as 3rd place on Task 2. 
\end{abstract}



\maketitle

\section{Introduction}
Retrieval augmented generation (RAG) has been a popular approach for question answering systems for some time \cite{rag_paper}, although recently has become a very popular approach for a wide range of tasks due to the zero-shot capabilities of large language models (LLMs) with an appropriate prompt and access to the relevant context for the task. Despite the existence of numerous question answering benchmarks, many do accurately reflect the diverse usage of current RAG systems. Thus, tracking both the efficacy of certain RAG architectures as well as tracking process remains difficult. The CRAG \cite{crag_paper} benchmark aims to resolve this with 3 different subtasks representing realistic RAG usage scenarios. 
The final key element of the CRAG benchmark is its scoring metric which explicitly punishes hallucinations. With the rising capabilities of LLMs, increasingly their outputs are taken at face value, despite the known issue of hallucinations. This has led to  high profile incidents causing concern with their use \cite{lawyer_chatgpt}. The CRAG score aims to punish hallucinated answers and encourages returning missing answers, equivalent to returning "i don't know" from the model, by giving scores of 1, 0, and -1 to correct, missing, and hallucinated answers respectively.
To address these various tasks, we train individual adapters for each of the various tasks, as well as API call generation required for accessing information in the mock API. This approach allows us to use a single LLama 3 \cite{llama3_paper} in memory while swapping out adapters based on the current needs. 

\section{Related Works}
The initial approach of Lewis et al. \cite{rag_paper} showed the benefits of providing additional text context for seq2seq models for NLP tasks that are knowledge intensive. Using BART, they were able to improve question answering tasks using dual biencoders for retrieval and training the model jointly, without the need for knowing which documents were relevant.

Adapters have become increasingly used since introduced by Houlsby et al. \cite{adapters}. LoRa \cite{lora_paper} has become a popular adapter approach, particularly for LLMs as they have grown substantially larger in recent years. The use of adapters allows modifying a model's output without training the entire network, which substantially saves on VRAM memory when training. Hu et al. \cite{lora_paper} discovered that when replacing a dot product between large vectors with an intermediate dot product of a much lower rank vector, the impacts on performance were minimal while further reducing the training parameters required. 

Finally, Stickland and Murray \cite{bert_pals} produced a multi-adapter model based on BERT, an approach that our system follows. In particular for the GLUE \cite{glue_paper} benchmark, which is comprised of multiple datasets, they showed that simply training a task specific adapter per dataset, they could improve the average performance by 0.5 points for BERT, while only introducing 10\% more parameters.
\begin{figure*}[t] 
    \centering
    \includegraphics[width=\textwidth]{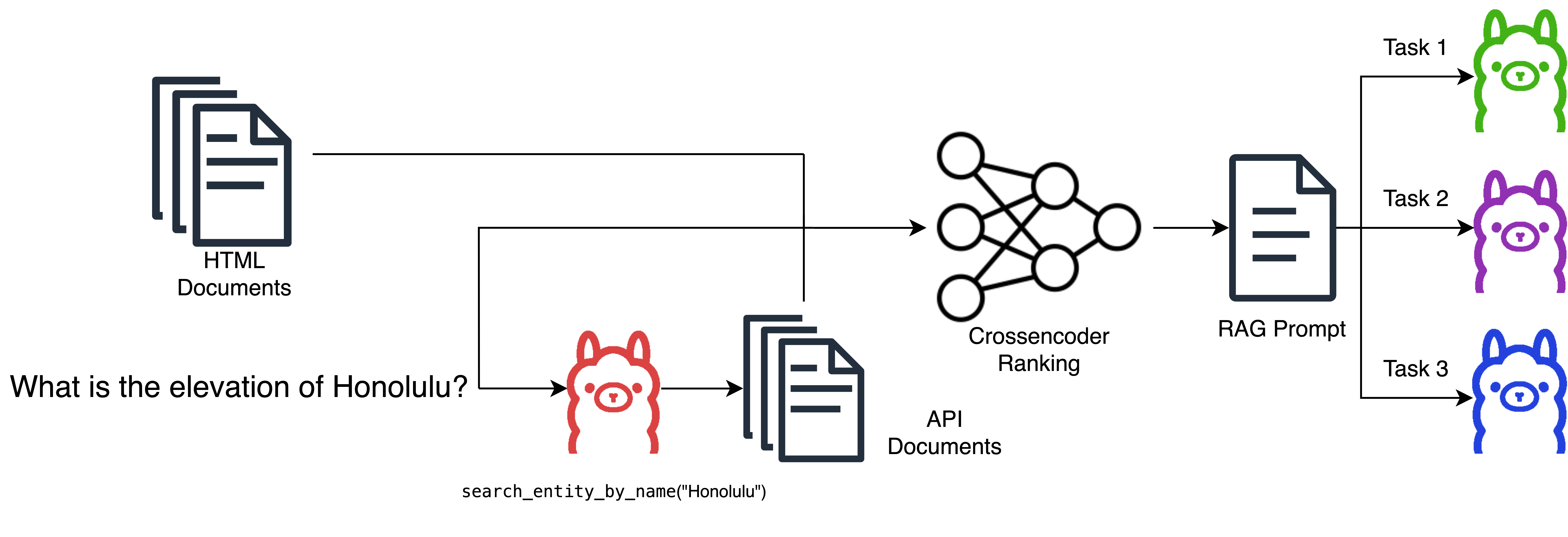}
    \caption{Pipeline for MARAGS. Each Llama in the figure represents a distinct LoRa model that can be rapidly swapped out during inference and is trained for its specific task. These tasks include the API call generation and the final question answering, for each task. Note, Task 1 does not include API documents in its final prompt.}
    \label{fig:maragas_figure}
\end{figure*}
\section{CRAG Dataset}
CRAG is a question answering dataset aimed at providing a realistic task to benchmark RAG systems as they are used in practice with a diverse set of questions, including 8 distinct question types and 5 distinct domains. Additionally, two sources of diversity which pose difficulty for LLMs are how dynamic a question's answer is and the popularity of the topic of the question. As shown in the baseline systems, real-time answers pose a challenge for RAG systems and they similarly struggle when the topic of the question is less common (referred to as "torso" and "tail" questions).

\subsection{Task 1}
For the first task the system must process 5 candidate HTML documents for generating answers, reflecting a standard web-based RAG application. A caveat is that the 5 candidates are sampled from the top-10 relevant documents retrieved from web search. Thus, there is no guarantee that the relevant information for answering the question is actually found within the top 5 documents. This creates an interesting challenge for hallucinations, as in some cases the answer should be easily generated by the model without the context from retrieved documents.
\subsection{Task 2}
Task 2 reflects a more sophisticated RAG scenario, where the system is provided with the same 5 HTML documents from before, however it now has access to a knowledge graph, accessible via a REST API. The system must determine which API endpoint to call, with the correct arguments, to retrieve additional relevant context from the knowledge graph. 
\subsection{Task 3}
Finally, Task 3 represents an extension of Task 2, where the system has access to both HTML and the knowledge graph API. However, in this task, the number of HTML documents to be processed is 50. This task is meant to measure both the computational efficiency of the approach as well as its ability to filter large amounts of potentially irrelevant information. 
\section{MARAGS Pipeline}
\subsection{Webpage Processing}
Our webpage processing pipeline utilizes BeautifulSoup4 \cite{richardson2007beautiful} for generating candidate segments from the HTML documents provided to the system. A common difficulty with RAG systems is determining a process for segmenting documents into smaller chunks to narrow the candidates to relevant sections and also reducing the length of the text sent to the model, given that a single document could exceed the context window of a model. 

In our case, we utilize the structure of the HTML to provide the segmentation. We traverse the tree structure of the parse HTML with breadth-first search. Any time the length of the text contained within a node (which includes all of the text of its descendants) is less than 2000 characters, we treat that as a segment. If a leaf node is reached with greater than 2000 characters, the text is split on the white space and into as many segments are needed such that each segment is under the threshold. The segment length was determined via inspection of HTML documents and their associated segmentation, thus future work could treat this as a hyperparameter and tune for performance.
\subsection{API Call Generation}
For Task 2 and 3, the knowledge graph mock API is available to be used for gathering additional information. The difficulty, however, is not only determining which API endpoint is the most appropriate, but also the proper arguments and their formatting for getting valid results from the API. 

Each API endpoint was transformed to a Python function with relevant documentation describing the purpose of the endpoint, the arguments, and what the endpoint returned. Each function also has an associated formatting function, which takes the returned JSON and converts it into segmented strings. The doc strings for each Python function are used to provide additional information to help guide the model on which one is the most appropriate to use.

For training a model to generate API calls with associated arguments, we use LoRa \cite{lora_paper} to train one of the several adapters we use with Llama 3 8B. For generating the target string for training, we first use Llama 3 to generate an initial prediction for the API call. Any initial prediction that successfully calls a function is retained as a target, regardless of whether or not the relevant information for the question is contained in the returned JSON. Initial predictions that fail to make a successful call are inspected and manually corrected if the correct function call is clear from the initial prediction and the question. Again, the manually modified targets are evaluated only on successfully calling an endpoint, though not validating that the relevant information is returned by the API. Any question where the target cannot be quickly modified is changed to a target of "None". 

We acknowledge that this approach to annotation is not optimal, as it likely results in successful, but incorrectly selected API endpoint calls. However, manually annotating each question to determine the correct API call and validating the returned information were indeed relevant would have been too time consuming given the size of the dataset.
\subsection{Candidate Ranking}
For candidate ranking, we attempted 4 different candidate ranking approaches. We utilized TF-IDF, a biencoder, cross-encoder, and an ensemble of the mentioned approaches (using mean rank as the ranking metric). Our TF-IDF implementation is based on Scikit-Learn \cite{scikit-learn}. The biencoder and cross-encoder are from the SentenceTransformer \cite{reimers-2019-sentence-bert} library, specifically the "multi-qa-MiniLM-L6-cos-v1" \footnote{https://huggingface.co/sentence-transformers/multi-qa-MiniLM-L6-cos-v1} and "ms-marco-MiniLM-L-6-v2" \footnote{https://huggingface.co/cross-encoder/ms-marco-MiniLM-L-6-v2} respectively.  

Evaluating candidate ranking in isolation is difficult, as relevant information is not labeled, so using the system accuracy and CRAG score is the most straight forward way to compare differences in each system. However, to test the various systems, we use the base Llama-3 8B model with no adapters for each retrieval approach and use the accuracy metric to determine the best performing approach. We use accuracy instead of CRAG at this stage for ranking, as we think this is a better representation of how often the relevant information retrieved. For a test set, we randomly select 500 samples from the Task 1 dataset.

\begin{table}[t]
\centering
\caption{A comparison of the retrieval approaches on a 500 set sample of CRAG dev set against Accuracy and CRAG Score.}
\label{tab:retrieval_scores}
\begin{tabular}{|l|c|c|}
\hline
Retrieval Model & Accuracy & CRAG \\
\hline
TF-IDF & 0.2740 & \textbf{-0.110} \\
\hline
Biencoder & 0.310 & -0.132 \\
\hline
Cross-encoder & \textbf{0.328} & -0.116 \\
\hline
Ensemble (mean rank) & 0.308 & -0.128 \\
\hline
\end{tabular}
\end{table}

The results of this experiment are shown in Table \ref{tab:retrieval_scores}. From the results, the cross-encoder is the best performing system, thus we used it for our retriever. We suspect that with proper tuning TF-IDF and ensembling would be much more performant overall, but as mentioned running extensive experimentation is difficult as it requires LLM generation to get an overall accuracy score. Using an LLM to label passages as relevant or not is a possible approach to allow for tuning of just the retriever, however we did not explore this.

Despite the cross-encoder being the most computationally expensive approach, we found it to be fast enough for processing the candidates in the required 30 seconds per sample. In the case of Task 3, it was necessary to use Python's multiprocessing library to process multiple HTMLs simultaneously to meet the runtime requirement.
\subsection{Retrieval Augmented Generation}
Finally, with ranked candidates, we use Llama 3 8B to augment generation with the relevant context for the question. We ran experiments with 2 different prompt structures, the primary difference between them being the ordering of the question and context. 

Our initial prompt structure started with the question first, then all of the retrieved candidates prior to the Llama model response, however we noticed that often due to how much context would be provided, the model would occasionally forget the question being asked. For example a question like "What is the elevation of Honolulu?" would result in an answer of "Honolulu population is 343,421", indicating the model remembered the subject of the question, but incorrectly answered with the population, rather than elevation. Our subsequent prompt structure placed the question after the context, which resolved the issue.

For training Llama 3, we trained LoRa models for each task individually. Given the penalization of hallucinations in the scoring metric, we try to take steps to mitigate further hallucinations due to fine-tuning, as it has been observed that fine-tuning LLMs can be a source of further hallucinations \cite{ft_hallucinations}. This likely applies to RAG systems in cases where the expected answer is not answerable given the context provided, i.e. no relevant information is given and the question is not answerable without further context, yet the model is trained to output the target answer regardless. Thus for our training setup, we first relabel the target for training samples in cases where our candidate retrieval system likely does not provide the correct information and Llama does not know the answer without that information. We use the provided dev set for training, with the 500 set sample used for retrieval comparison treated as our holdout set. 

\begin{table}[t]
\centering
\caption{A comparison across inference models test. Training on relabeled "hittable" targets hurts accuracy, but provides the best CRAG Score overall.}
\label{tab:crag_scores}
\begin{tabular}{|l|c|c|c|}
\hline
Model & Accuracy& Hallucination & CRAG \\
\hline
Llama 3 8B & 0.328 & 0.4440 & -0.1160 \\
\hline
 - LoRa & \textbf{0.398} & 0.602 & -0.204 \\
\hline
 - LoRa (relabeled) & 0.242 & \textbf{0.056} & \textbf{0.186} \\
\hline
\end{tabular}
\end{table}

\begin{figure*}[t] 
    \centering
    \includegraphics[width=\textwidth]{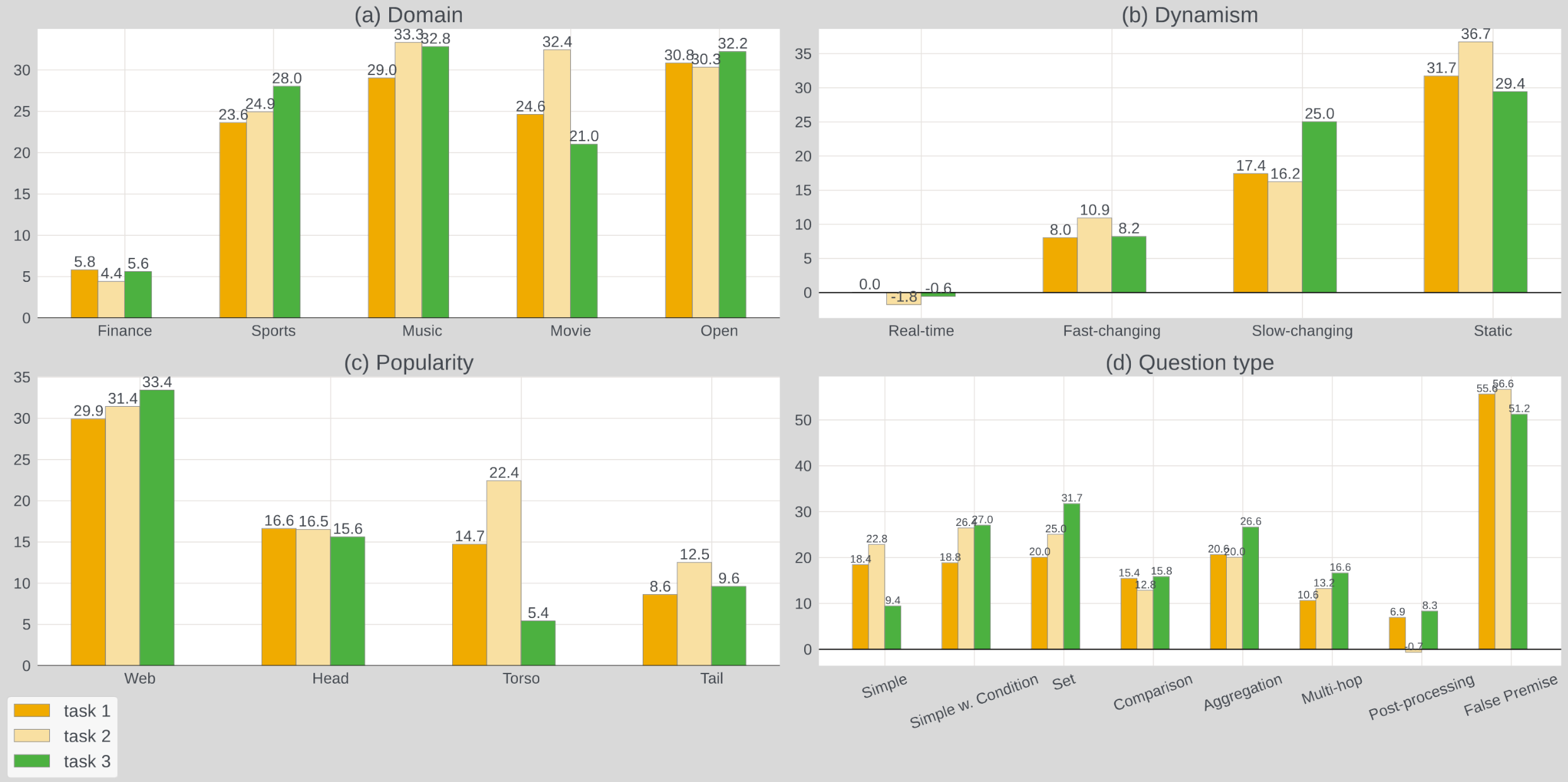}
    \caption{CRAG Score results on the test dataset calculated via manual assessment.}
    \label{fig:results}
\end{figure*}

Our initial approach for determining which training samples need relabeling has been explored previously \cite{noisey_labels}. A common and simple approach to filter/relabel incorrectly\footnote{"Incorrectly" here simply means in the context of our retrieval system, not that the provided answer is not true.} labeled samples is to use a particular sample's training loss after training to the point of over-fitting. High loss examples after over-fitting likely indicate examples that are incorrectly labeled and thus can be filtered out. Not all samples with high loss will be incorrect labels, instead simply being hard examples, yet typically the benefit of disposing of incorrectly labeled samples outweighs ignoring difficult ones.

Initial experiments, however, indicated that this method did not work well on finance based questions. Further analysis would be required for a more definitive answer, though we suspect that this is due to the fact that the loss in hallucinated answers when dealing with numeric outputs is likely less than typical string outputs. For example, with a question "What was Apple's closing price today?", with a hypothetical correct answer "\$203.51", a prediction of "\$204.52" would likely not result in filtering via this method. Compare that with a question such as "Which movie won Best Picture in 1990?" with an answer of "Driving Miss Daisy" and a prediction of "Dancing with Wolves", the loss will be comparatively much higher.

We instead determine these samples by first running the system with the base Llama 3 model, with a prompt indicating to always produce a prediction for each of the 4 candidate retrieval approaches mentioned previously. We use GPT-4o to evaluate whether any of the generated answers are correct. If any are correct, the original label is retained for the training, otherwise "i don't know" is used as the target label. In the case of false premise questions, we always retain the original label as the target label. We repeat this process for each task, given that each has access to different data sources, to generate a training dataset for each LoRa adapter. 

We use the llama-recipes repository \cite{llama_recipes} for training the LoRa adapters, utilizing the default LoRa configuration. The only modifications were changing the LoRa rank from 8 to 256 and increasing the weight decay from 0.01 to 1.0. 

We demonstrate the effectiveness of relabeling in Table \ref{tab:crag_scores}. We ran 3 different answer generation setups for the 500 sample Task 1 hold out set we created. The first is an unmodified Llama 3 8B model, the second is a LoRa model using the original targets, and the final a LoRa model with relabeled targets. As shown, using the original targets provides the best accuracy, but also worsens hallucinations over the base model. While using the relabeled targets hurts accuracy, it also substantially reduces hallucinations, providing the best CRAG Score among the three.

\section{Results}
As part of the competition, a manual evaluation was conducted on user submissions. The automatic evaluation throughout was dependent on scoring via GPT-3.5 Turbo, given that correct user submissions may not have been exact matches to the expected answer. However, issues such as prompt injection still pose problems for automatic evaluation via LLMs. The results of our system across the various aspects of the dataset are shown in Figure \ref{fig:results}. As can be seen by the results, our system suffers many of the problems the dataset is meant to expose with most RAG systems. 

Similar to the baseline systems for CRAG, finance was the most challenging domain. The exact reason warrants further analysis, though contributing factors likely include LLMs known issues with number processing and simple mathematics and the fact that much online finance data is often not stored in plain text, but rather visualizations such as graphs.

Dynamism proves to be the most challenging question categorization, with model performance steadily decreasing as a question becomes more dynamic. Real-time questions prove to be the most challenging question category of any of the breakouts. Our prompt structure did not include any meta-data provided by the HTML documents, such as publish data or access date, which likely would have improved performance on dynamic questions, although likely not significantly.

The performance difference between head, torso, and tail questions appeared less substantial than our original expectations, though clearly performance drops off as popularity falls. Interestingly, Task 3 underperforms the other tasks in head, torso, and tail. We suspect that including substantially more search results includes overlapping entities / subjects, at which point conflicting information would be difficult to resolve.

Finally, the most interesting results in the question type results are the false premise category. Our system was able to achieve scores similar to the SOTA systems featured in the CRAG paper, despite obviously being a much smaller system overall. Interestingly, the false premise questions were the only type where our training setup always kept the original target label, rather than mapping the target to "i don't know".

\section{Future Work}
Observations we had during the competition were instances of catastrophic forgetting due to our attempts to reduce hallucinations. For instance, the question "Who has had a longer musical career, Sharika or Machine Gun Kelly?" is answerable by Llama without context, simply based on the knowledge it has of the two artists. However, after LoRa training, questions like this and others were often answered with "i don't know" in cases where the answer was not discoverable in the retrieved information. Methods to prevent this is something we are interested in pursuing in future work. 

Additionally, we hope to explore larger Llama models, 70B+, in the future for this task. We were unable to get the 70B model running in the competition compute environments, so did not spend much time looking at larger models. However, it is very likely moving to larger models would provide a substantial improvement over the 8B model.

\section{Conclusion}
In this work we presented MARAGS, a multi-adapter solution to the CRAG dataset. We demonstrated the effectiveness of training individual LoRa adapters for the 4 tasks in the pipeline, specifically API call generation and Task 1,2, and 3 answer generation. CRAG presents a variety of different tasks and questions to allow the tracking the progress of various methods used to build RAG systems. The penalization of hallucinations is a unique and important feature as future AI systems become increasingly common throughout society, as hallucinations hurt user trust in these systems. We discussed our methods for reducing these hallucinations, but they are not without cost, as in some cases the model fails to output previously known knowledge. Clearly the importance of balancing these two factors is a key to leveraging LLMs to their full potential, while also improving user trust.

\begin{acks}
We are grateful to the KDD Cup organizers, Meta, and AIcrowd for all the work that goes into hosting a successful competition.  
\end{acks}

\bibliographystyle{ACM-Reference-Format}
\bibliography{sample-base}

\end{document}